%% file: 5913.tex
\documentclass[runningheads]{llncs}
\usepackage{graphicx}
\usepackage{comment}
\usepackage{amsmath,amssymb,subfigure} 
\usepackage{color}
\usepackage{makecell}
\usepackage{adjustbox}  
\usepackage{multirow}
\usepackage{xpunctuate} 
\usepackage{foreign}
\usepackage{pbox}
\usepackage{tabularx,booktabs,calc}
\usepackage{hhline}
\usepackage{enumitem}
\usepackage{stackengine}
\usepackage{url}
\usepackage{hyperref}

\begin{document}
\pagestyle{headings}
\mainmatter
\def\ECCVSubNumber{5913}

\title{3D-CVF: Generating Joint Camera and
LiDAR Features Using Cross-View Spatial Feature Fusion for 3D Object Detection}

 
\titlerunning{3D-CVF}

\authorrunning{Jin Hyeok Yoo et al.}
\author{Jin Hyeok Yoo\thanks{: Equal contribution}\and
Yecheol Kim$^*$\and
Jisong Kim\and
Jun Won Choi\thanks{: Corresponding author}}

\institute{Department of Electrical Engineering, Hanyang University\\
\email{\{jhyoo,yckim,jskim\}@spa.hanyang.ac.kr}\\
\email{junwchoi@hanyang.ac.kr}}

\maketitle

\begin{abstract}
In this paper, we propose a new deep architecture for fusing camera and LiDAR sensors for 3D object detection. Because the camera and LiDAR sensor signals have different characteristics and distributions, fusing these two modalities is expected to improve both the accuracy and robustness of 3D object detection. One of the challenges presented by the fusion of cameras and LiDAR is that the spatial feature maps obtained from each modality are represented by significantly different views in the camera and world coordinates; hence, it is not an easy task to combine two heterogeneous feature maps without loss of information. To address this problem, we propose a method called 3D-CVF that combines the camera and LiDAR features using the cross-view spatial feature fusion strategy. First, the method employs {\it auto-calibrated projection}, to transform the 2D camera features to a smooth spatial feature map with the highest correspondence to the LiDAR features in the bird's eye view (BEV) domain. Then, a {\it gated feature fusion network} is applied to use the spatial attention maps to mix the camera and LiDAR features appropriately according to the region. Next, camera-LiDAR feature fusion is also achieved in the subsequent proposal refinement stage.
The low-level LiDAR features and camera features are separately pooled using {\it region of interest (RoI)-based feature pooling} and fused with the joint camera-LiDAR features for enhanced proposal refinement. Our evaluation, conducted on the KITTI and nuScenes 3D object detection datasets, demonstrates that the camera-LiDAR fusion offers significant performance gain over the LiDAR-only baseline and that the proposed 3D-CVF achieves state-of-the-art performance in the KITTI benchmark.
\end{abstract}

\section{Introduction}

\hspace{\parindent}Object detection has been considered one of the most challenging computer vision problems. Recently, the emergence of convolutional neural networks (CNN) has enabled unprecedented progress in object detection techniques owing to its ability to  extract the abstract high-level features from the 2D image.
Thus far, numerous object detection methods have been developed for 2D object detection \cite{ssd,yolo,fasterrcnn}. Recently, these studies have been extended to the 3D object detection task \cite{pixor,mv3d,avod,roarnet,fpointrcnn,fpointnet,voxelnet,second,mmf,pointpillar,contfuse,std,fconvnet}, where the locations of the objects should be identified in 3D world coordinates. 3D object detection is particularly useful for autonomous driving  applications because diverse types of dynamic objects, such as surrounding vehicles, pedestrians, and cyclists, must be identified in the 3D environment.
 \begin{figure*}[t]
    \centering
    \begin{subfigure}[]{\includegraphics[width=0.25\textwidth]{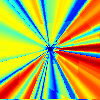}}
    \end{subfigure}
    \hspace{0cm}
    \begin{subfigure}[]{\includegraphics[width=0.25\textwidth]{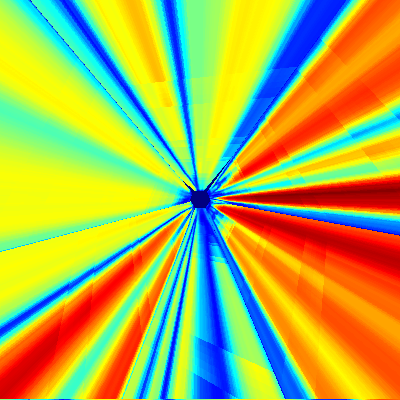}}
    \end{subfigure}
    \hspace{0cm}
    \begin{subfigure}[]{\includegraphics[width=0.25\textwidth]{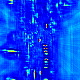}}
    \end{subfigure}
    \caption {{\bf Visualization of the projected camera feature map:} (a), (b), and (c) show visualizations of the six camera feature maps projected in the bird's eye view (BEV) domain.  Without our auto-calibrated projection, some artifacts in the feature map are visible in (a). The auto-calibrated projection generates the smooth and dense transformed feature map shown in (b). However, the feature map in (b) fails to localize the region of the objects. After applying the adaptive gated fusion network, we can finally resolve the region of objects as shown in the feature map (c).}
    \label{fig:interp_comp}
\end{figure*}

In general, achieving good accuracy in 3D object detection using only a camera sensor is not an easy task owing to the lack of depth information. Thus, other ranging sensors such as LiDAR, Radar, and RGB-D camera sensors are widely used as alternative signal sources for 3D object detection. 
Thus far, various 3D object detectors employing LiDAR sensors have been proposed, including MV3D \cite{mv3d}, PIXOR \cite{pixor}, ContFuse \cite{contfuse}, PointRCNN \cite{pointrcnn}, F-ConvNet \cite{fconvnet}, STD \cite{std}, VoxelNet \cite{voxelnet}, SECOND \cite{second}, MMF \cite{mmf}, PointPillar \cite{pointpillar}, and Part A$^2$ \cite{parta2}.
Although the performance of the LiDAR only based 3D object detectors have been significantly improved lately, LiDAR point clouds are still limited for providing dense and rich information on the objects such as their fine-grained shape, colors, and textures. Hence, using camera and LiDAR data together is expected to yield better and more robust detection results in accuracy. Various camera and LiDAR fusion strategies have been proposed for 3D object detection. 
Well-known camera and LiDAR fusion methods include AVOD \cite{avod}, MV3D \cite{mv3d}, MMF \cite{mmf}, RoarNet \cite{roarnet}, F-PointNet \cite{fpointnet}, and ContFuse \cite{contfuse}. 

In fact, the problem of fusing camera and LiDAR sensors is challenging as the features obtained from the camera image and LiDAR point cloud are represented in different points of view (i.e., camera-view  versus 3D world view).  When the camera feature is projected into 3D world coordinates, some useful spatial information about the objects might be lost since this transformation is a one-to-many mapping. Furthermore, there might be some inconsistency between the projected coordinate and LiDAR 3D coordinate. Indeed, it has been difficult for the camera-LiDAR fusion-based methods to beat the LiDAR-only methods in terms of performance.  This motivates us to find an effective way to fuse two feature maps in different views without losing important information for 3D object detection.

In this paper, we propose a new 3D object detection method, named {\it 3D-cross view fusion (3D-CVF)}, which can fuse the spatial feature maps separately extracted from the camera and LiDAR data, effectively. As shown in Fig.~\ref{fig:overall}, we are interested in fusing the LiDAR sensor and the $N$ multi-view cameras deployed to cover a wider field of view. Information fusion between the camera and LiDAR is achieved over two object detection stages.
In the first stage, we aim to generate the strong joint camera-LiDAR features.  
The {\it auto-calibrated feature projection} maps the camera-view features to smooth and dense BEV feature maps using the interpolated projection capable of correcting the spatial offsets. 
Fig. \ref{fig:interp_comp} (a) and (b) compare the feature maps obtained without auto-calibrated projection versus with the auto-calibrated projection, respectively. 
Note that the auto-calibrated projection yields a smooth camera feature map in the BEV domain as shown in Fig. \ref{fig:interp_comp} (b).  We also note from Fig. \ref{fig:interp_comp} (b) that since the camera feature mapping is a one-to-many mapping, we cannot localize the objects on the transformed camera feature
 map. 
To resolve objects in the BEV domain, we employ the {\it adaptive gated fusion network} that determines where and what should be brought from two sources using attention mechanism. 
Fig. \ref{fig:interp_comp} (c) shows the appropriately-localized activation for the objects obtained by applying the adaptive gated fusion network.
Camera-LiDAR information fusion is also achieved at the second proposal refinement stage. Once the region proposals are found based on the joint camera-LiDAR feature map obtained in the first stage, {\it 3D region of interest (RoI)-based pooling} is applied to fuse low-level LiDAR and camera features with the joint camera-LiDAR feature map. The LiDAR and camera features corresponding to the 3D RoI boxes are pooled and encoded by PointNet encoder. Aggregation of the encoded features with the joint camera-LiDAR features lead to improved proposal refinement.

We have evaluated our 3D-CVF method on publicly available KITTI \cite{kitti} and nuScenes \cite{nuScenes} datasets. We confirm that by combining the above two sensor fusion strategies combined, the proposed method offers up to 1.57\% and 2.74\% performance gains in mAP over the baseline without sensor fusion on the KITTI and nuScenes datasets,  respectively. Also, we show that the proposed 3D-CVF method achieves impressive detection accuracy comparable to state-of-the-art performance in KITTI 3D object detection benchmark.

The contributions of our work are summarized as follows
\begin{itemize}
    \item We propose a new 3D object detection architecture that effectively combines information provided by both camera and LiDAR sensors in two detection stages.  In the first stage, the strong joint camera-LiDAR feature is generated by applying the auto-calibrated projection and the gated attention.   In the second proposal refinement stage, 3D RoI-based feature aggregation is performed to achieve further improvements through sensor fusion. 
  
    \item We investigate the benefit of the sensor fusion achieved by the 3D-CVF. Our experiments demonstrate that the performance gain achieved by the sensor fusion in nuScenes dataset is higher than that in KITTI dataset. Because the resolution of LiDAR used in nuScenes is lower than that in KITTI, this shows that the camera sensor compensates low resolution of the LiDAR data. Also, we observe that the performance gain achieved by the sensor fusion is much higher for distant objects than for near objects, which also validates our conclusion. 
    
\end{itemize}

\section{Related Work}
\subsection{LiDAR-Only 3D Object Detection}
\hspace{\parindent}The LiDAR-based 3D object detectors should encode the point clouds since they have unordered and irregular structures.  MV3D \cite{mv3d} and PIXOR \cite{pixor} projected 3D point clouds onto the discrete grid structure in 2D planes and extracted the features from the resulting multi-view 2D images. PointRCNN \cite{pointrcnn} and STD \cite{std} used PointNet \cite{pointnet,pointnet++} to yield the global feature representing the geometric structure of the entire point set.
Voxel-based point encoding methods used 3D voxels to organize the unordered point clouds and encoded the points in each voxel using the point encoding network \cite{voxelnet}.
Various voxel-based 3D object detectors have been proposed, including SECOND \cite{second}, PointPillar \cite{pointpillar}, and Part-${A}^{2}$ \cite{parta2}.

\subsection{LiDAR and Camera Fusion-based 3D Object Detection}

\hspace{\parindent}To exploit the advantages of the camera and LiDAR sensors, various camera and LiDAR fusion methods have been proposed for 3D object detection. The approaches proposed in \cite{fpointnet,roarnet,pointfusion,fconvnet} detected the objects in the two sequential steps, where 1) the region proposals were generated based on the camera image, and then 2) the LiDAR points in the region of interest were processed to detect the objects. However, the performance of these methods is limited by the accuracy of the camera-based detector. MV3D \cite{mv3d} proposed the two-stage detector, where 3D proposals are found from the LiDAR point clouds projected in BEV, and 3D object detection is performed by fusing the multi-view features obtained by RoI pooling. AVOD \cite{avod} fused the LiDAR BEV and camera front-view features at the intermediate convolutional layer to propose 3D bounding boxes. ContFuse \cite{contfuse} proposed the effective fusion architecture that transforms the front camera-view features into those in BEV through some interpolation network. 
MMF \cite{mmf} learned to fuse both camera and LiDAR data through multi-task loss associated with 2D and 3D object detection, ground estimation, and depth completion. 

While various sensor fusion networks have been proposed, they do not easily outperform LiDAR-only based detectors.
This might be due to the difficulty of combining the camera and LiDAR features represented in different view domains.
In the next sections, we present an effective way to overcome this challenge. 

\section{Proposed 3D Object Detector}
\hspace{\parindent}In this section, we present the details of the proposed architecture.

\begin{figure*}[t]
    \centering
    \centerline{\includegraphics[width=1.0\textwidth]{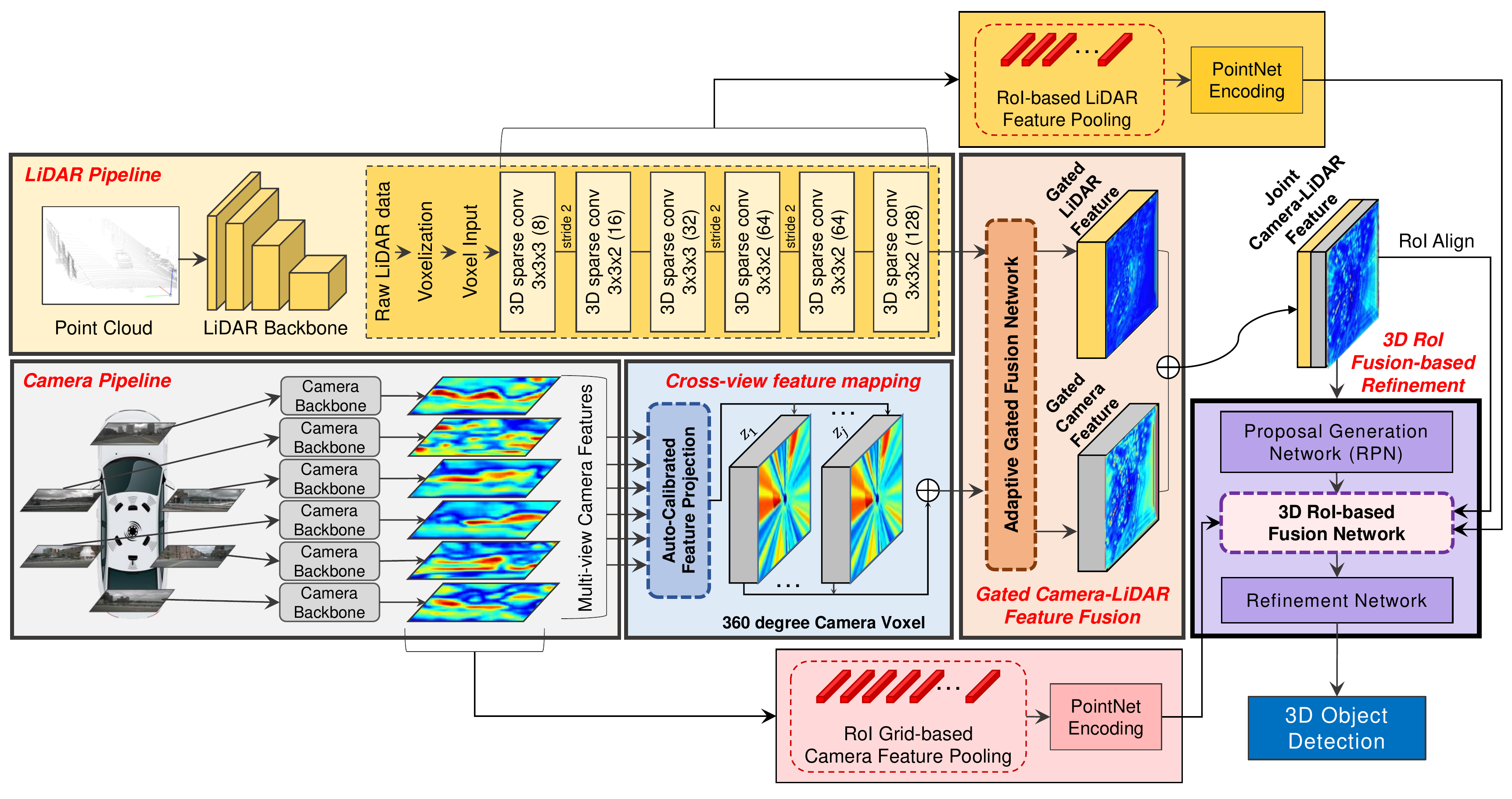}}
    \caption {{\bf Overall structure of 3D-CVF: } 
    After point clouds and each camera-view image are separately processed by each backbone network, the camera-view features are transformed to the features in BEV using the auto-calibrated feature projection. Then, the camera and LiDAR features are fused using the gated feature fusion network. The detection outputs are predicted after refining the proposals using 3D RoI-based fusion network. The format of 3D convolutional layers used in the figure follows ``$k_x$ x $k_y$ x $k_z$ (channel size)" where $k_x$, $k_y$ and $k_z$ denote the kernel sizes in each axis.}
    \label{fig:overall}
\end{figure*}

\subsection{Overall architecture}
\hspace{\parindent}The overall architecture of the proposed method is illustrated in Fig. \ref{fig:overall}. It consists of five modules including the 1) LiDAR pipeline, 2) camera pipeline, 3) cross-view spatial feature mapping, 4) gated camera-LiDAR feature fusion network, and 5) proposal generation and refinement network. Each of them is described in the following 

{\bf LiDAR Pipeline:} LiDAR points are first organized based on the LiDAR voxel structure. The LiDAR points in each voxel are encoded by the point encoding network \cite{voxelnet}, which generates the fixed-length embedding vector. These encoded LiDAR voxels are processed by six 3D sparse convolution \cite{second} layers with  stride two, which produces the LiDAR feature map of 128 channels in the BEV domain. After sparse convolutional layers are applied, the 
width and height of the resulting LiDAR feature map are reduced by a factor of eight compared to those of the LiDAR voxel structure.

{\bf RGB Pipeline:} In parallel to the LiDAR pipeline, the camera RGB images are processed by the CNN backbone network. We use the pre-trained ResNet-18 \cite{resnet} followed by feature pyramid network (FPN) \cite{fpn} to generate the camera feature map of 256 channels represented in camera-view. The width and height of the camera feature maps are reduced by a factor of eight compared to those of the input RGB images.

{\bf Cross-View Feature Mapping:} The cross-view feature (CVF) mapping generates the camera feature maps projected in BEV. The auto-calibrated projection converts the camera feature maps in camera-view to those in BEV. Then, the projected feature map is enhanced by the additional convolutional layers and delivered to the gated camera-LiDAR feature fusion block. 

{\bf Gated Camera-LiDAR Feature Fusion:} The adaptive gated fusion network is used to combine the camera feature maps and the LiDAR feature map. The spatial attention maps are applied to both feature maps to adjust the contributions from each modality depending on their importance. The adaptive gated fusion network produces the joint camera-LiDAR feature map, which is delivered to the 3D RoI fusion-based refinement block. 

{\bf 3D RoI Fusion-based Refinement: } 
After the region proposals are generated based on the joint camera-LiDAR feature map, the RoI pooling is applied for proposal refinement. Since the joint camera-LiDAR feature map does not contain sufficient spatial information, both the multi-scale LiDAR features and camera features are extracted using 3D RoI-based pooling. These features are separately encoded by the PointNet encoder and fused with the joint camera-LiDAR feature map by a 3D RoI-based fusion network. The fused feature is finally used to produce the final detection results. 

\begin{figure*}[t]
    \centering
    \centerline{\includegraphics[width=0.8\textwidth]{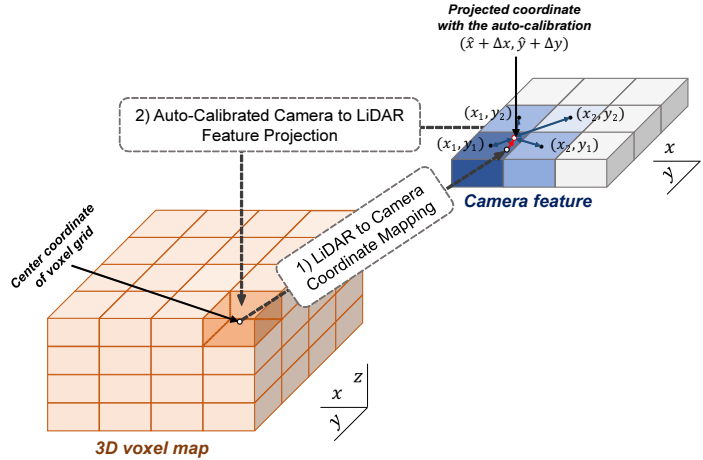}}
    \caption {{\bf Illustration of the proposed auto-calibrated projection: }To represent the camera feature in BEV, the center coordinate of a voxel is projected onto the point $(\hat{x},\hat{y})$ with calibration offset $(\Delta{x},\Delta{y})$ in the camera-view plane. The neighboring four feature pixels are combined using linear interpolation and assigned to the corresponding voxel.}
    \label{fig:projection}
\end{figure*}
\subsection{Cross-View Feature Mapping}

\hspace{\parindent} {\bf Dense Camera Voxel Structure:} The camera voxel structure is used for the feature mapping. To generate the spatially dense features, we construct the camera voxel structure whose width and height are two times longer than those of the LiDAR voxel structure in the $(x,y)$ axis. This leads to the voxel structure with higher spatial resolution. In our design, the camera voxel structure has four times as many voxels as the LiDAR voxel structure. 

{\bf Auto-Calibrated Projection Method:} 
The auto-calibrated projection technique is devised to 1) transform the camera-view feature into the BEV feature and 2) find the best correspondence between them to maximize the effect of information fusion. The structure of the auto-calibrated projection method is depicted in  Fig. \ref{fig:projection}. First, the center of each voxel is projected to $(\hat{x},\hat{y})$ in the camera-view plane using the world-to-camera-view projection matrix and 
$(\hat{x},\hat{y})$ is adjusted by the calibration offset $(\Delta{x},\Delta{y})$.
Then, the neighbor camera feature pixels near to the calibrated position $(\hat{x}+\Delta{x},\hat{y}+\Delta{y})$ are combined with the weights determined by interpolation methods. That is, the combined pixel vector $\mathbf{u}$ is given by
\begin{gather}
\mathbf{u} = \sum_{m=1}^{2}\sum_{n=1}^2 w_{m,n} \mathbf{f}_{m,n},
\end{gather}
where the set $\{\mathbf{f}_{m,n}\}$ corresponds to four adjacent feature pixels  closest to  $(\hat{x}+\Delta{x},\hat{y}+\Delta{y})$, and $w_{m,n}$ is the weight obtained by the interpolation methods. In bilinear interpolation, $w_{m,n}$ is obtained using Euclidean distance as follows
\begin{gather}
w_{m,n} \propto \left|(x_m,y_m)-(\hat{x}+\Delta{x},\hat{y}+\Delta{y}))\right|^{-1},
\end{gather}
where $w_{m,n} $ is normalized such that $\sum_{m=1}^{2}\sum_{n=1}^2 w_{m,n}=1$.
Then, the combined feature $\mathbf{u}$ is assigned to the corresponding voxel. Note that  different calibration offsets $(\Delta{x},\Delta{y})$ are assigned to different regions in 3D space. These calibration offset parameters can be jointly optimized along with other network weights. The auto-calibrated projection provides spatially smooth camera feature maps that best match with the LiDAR feature map in the BEV domain. 

\begin{figure}[t]
    \centering
    \centerline{\includegraphics[width=0.5\textwidth]{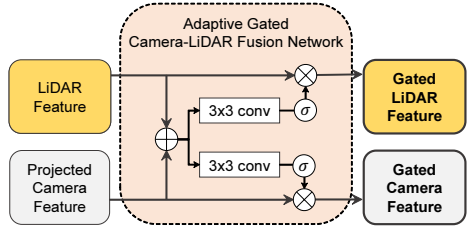}}
    \caption {{\bf Adaptive gated fusion network: } The adaptive gated fusion network generates the attention maps by applying  $3\times3$ convolutional layer followed by a sigmoid function to the concatenated inputs. These attention maps are multiplied to both camera and LiDAR features through the element-wise product operation.}
    \label{fig:gating}
\end{figure}

\subsection{Gated Camera-LiDAR Feature Fusion} 
\hspace{\parindent} {\bf Adaptive Gated Fusion Network:} To extract  essential features from both camera and LiDAR sensors, we apply an adaptive gated fusion network that selectively combines the feature maps depending on the relevance to the object detection task \cite{jkkim_accv}.  The proposed gated fusion structure is depicted in Fig. \ref{fig:gating}. The camera and LiDAR features are gated using the attention maps as follows
\begin{align}
\mathbf{F}_{g.C}&=\mathbf{F}_{C}\times\underbrace{\sigma(\text{Conv}_C(\mathbf{F}_{C}\oplus \mathbf{F}_{L}))}_{\text{Camera Attention Map}}\\
\mathbf{F}_{g.L}&=\mathbf{F}_{L}\times\underbrace{\sigma(\text{Conv}_L(\mathbf{F}_{C}\oplus \mathbf{F}_{L}))}_{\text{LiDAR Attention Map}}
\end{align}
where $\mathbf{F}_{C}$ and $\mathbf{F}_{L}$ represent the camera feature and LiDAR feature, respectively, $\mathbf{F}_{g.C}$ and $\mathbf{F}_{g.L}$ are the corresponding gated features, $\times$ is the element-wise product operation, and $\oplus$ is the channel-wise concatenation operation. 
Note that the elements of the attention maps indicate the relative importance of the camera and LiDAR features. After the attention maps are applied,  the final joint feature $\mathbf{F}_{joint}$ is obtained by concatenating $\mathbf{F}_{g.C}$ and $\mathbf{F}_{g.L}$ channel-wise.  (see Fig.~\ref{fig:overall}.) 

\subsection{3D-RoI Fusion-based Refinement}

\hspace{\parindent} {\bf Region Proposal Generation: }
The initial detection results are obtained by the region proposal network (RPN). Initial regression results and objectness scores are predicted by applying the detection sub-network to the joint camera-LiDAR feature. Since the initial detection results have a large number of proposal boxes associated with objectness scores, the boxes with high objectness scores remain through NMS post-processing with the IoU threshold 0.7.

\begin{figure*}[t]
    \centering
    \centerline{\includegraphics[width=0.93\textwidth]{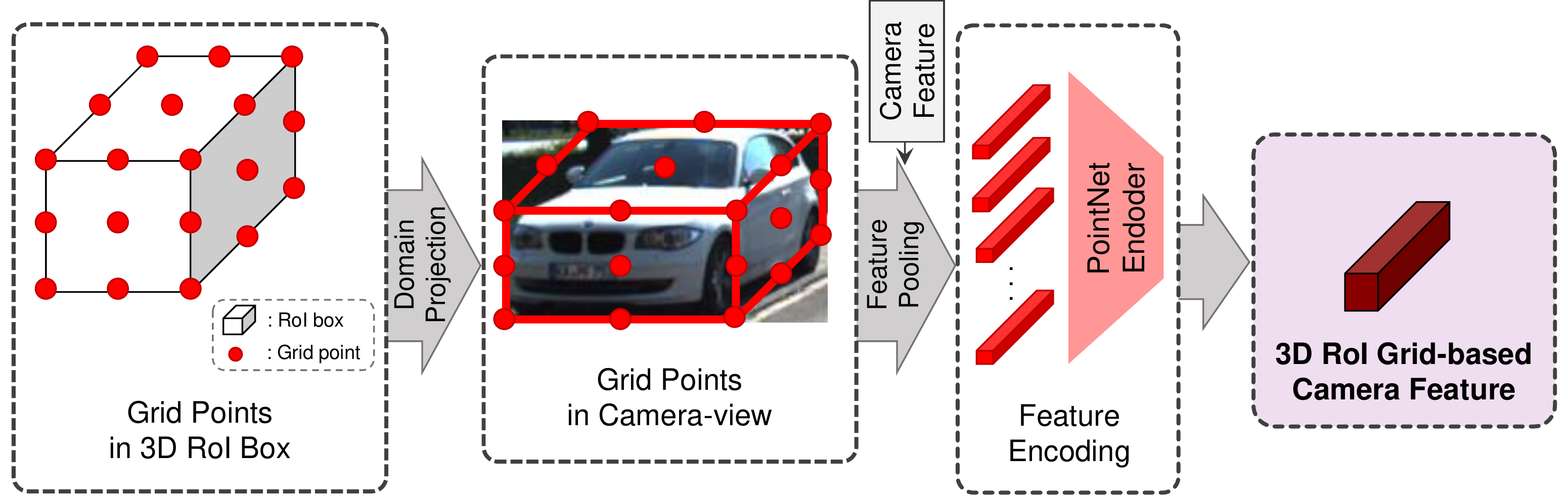}}
    \caption {{\bf Illustration of the proposed RoI grid-based pooling of camera features: } The RoI grid-based camera feature is generated by pooling the camera features according to the grid points in a 3D RoI box and encoding them using PointNet encoder. 
    }
    \label{fig:grid}
\end{figure*}
{\bf 3D RoI-based Feature Fusion: }
The predicted box regression values are translated to the global coordinates using the rotated 3D RoI alignment  \cite{mmf}. The low-level LiDAR and camera features are pooled using 3D RoI-based pooling and combined with the joint camera-LiDAR features. These low-level features retain the detailed spatial information on objects (particularly in $z$ axis) so that it can provide useful information for refining the region proposals. 
Specifically, six multi-scale LiDAR features corresponding to the 3D RoI boxes are pooled by 3D RoI-based pooling. These low-level LiDAR features are individually encoded by PointNet encoders for each scale and concatenated into a $1 \times 1$ feature vector. 
Simultaneously, the multi-view camera features are also transformed into a $1 \times 1$ feature vector. 
 Since the camera-view features are represented in a different domain from the 3D RoI boxes, we devise the {\it RoI grid-based pooling}. As shown in Fig. \ref{fig:grid}, consider the $r\times r\times r$ equally spaced coordinates in the 3D RoI box. These points are projected to the camera view-domain and the camera feature pixels corresponding to these points are encoded by the PointNet encoders. Concatenation of these encoded multi-view camera features forms another $1 \times 1$ feature vector.  The final feature used for proposal refinement is obtained by concatenating these two $1 \times 1$ feature vectors with the RoI aligned joint camera-LiDAR features.

\subsection{Training Loss Function}
\hspace{\parindent}Our 3D-CVF is trained via two-stage training process. In the first stage, we train the network pipeline up to RPN using the RPN loss, $L_{rpn} = \beta_1 L_{cls} + \beta_2 (L_{reg|\theta}+L_{reg|loc})$, where $\beta_1$ and $\beta_2$ are set to $1.0$ and $2.0$, respectively, and $L_{reg|loc}$ and $L_{reg|\theta}$ are given by the Smoothed-L1 loss \cite{fastrcnn} and modified Smoothed-L1 loss \cite{second}, respectively. Note that we follow suggestions from \cite{second} in parameterizing  3D ground truth boxes and 3D anchors. Note also that $L_{cls}$ denotes the focal loss \cite{focal} 
\begin{align}
 L_{cls} &= \frac{1}{N_{box}}\sum_{i=1}^{N_{box}}-\alpha(1-p_i)^\gamma\log(p_i),
\end{align}
where $N_{box}$ denotes the total number of boxes, $p_i$ is the objectness scores for $i$th box, and we set $\alpha=0.25$ and $\gamma=2$.  
In the next stage, the entire network is trained using the RPN loss $L_{rpn}$
plus refinement loss $L_{ref}$. The refinement loss $L_{ref}$ is given by 
\begin{align}
L_{ref} = \beta_1 L_{iou} + \beta_2(L_{reg|\theta}+ L_{reg|loc}),
\end{align}
where 
 $L_{iou}$ denotes the confidence score refinement loss that follows the definition of 3D IoU loss in \cite{Gs3d}.
Further details of training procedure are provided in the next section.

\section{Experiments}
\hspace{\parindent} In this section, we evaluate the performance of the proposed 3D-CVF on the KITTI \cite {kitti} and nuScenes \cite{nuScenes} datasets. 
\subsection{KITTI}
\hspace{\parindent}The KITTI dataset is the widely used dataset for evaluating 3D object detectors. It contains the camera and LiDAR data collected using a single Pointgrey camera and Velodyne HDL-64E LiDAR. The training set and test set contain 7,481 images and 7,518 images, respectively.
For validation, we split the labeled training set into the {\it train} set and {\it valid} set by half as done in \cite{mv3d}. The detection task is divided into  three different levels of difficulty, namely  ``easy", ``moderate", and ``hard". The average precision (AP) obtained from the 41-point precision-recall (PR) curve was used as a performance metric. 
\input{./table/test_3d.tex}

{\bf Training Configuration: } We limited the range of point cloud to [$0,70.4$] $\times$[$-40,40$]$\times$[$-3,1$]$m$ in $(x,y,z)$ axis. The LiDAR voxel structure consists of $1600\times1408\times40$ voxel grids with each voxel of size $0.05\times0.05\times0.1m$. We aimed to detect only cars, because the training data for other categories is not large enough in KITTI dataset. Accordingly, only two anchors with different angles (0$^{\circ}$, 90$^{\circ}$) were used. 
To train the 3D-CVF, we used the pre-trained LiDAR backbone network. As mentioned, training was conducted in two stages. We first trained the network up to RPN using the ADAM optimizer with one-cycle learning rate policy \cite{1cycle} over 70 epochs. The learning rate was scheduled with the max parameter set to 3e-3, the division factor 10, the momentum range from 0.95 to 0.85, and the fixed weight decay parameter of 1e-2. The mini-batch size was set to 12.  Next, the entire network was trained over 50 epochs with the mini-batch size of 6. The initial learning rate was set to 1e-4 for the first 30 epochs and decayed by a factor of 0.1 every 10 epochs. 
As a camera backbone network, we used the ResNet-18 \cite{resnet} network with FPN \cite{fpn} pre-trained with the KITTI 2D object detection dataset.

{\bf Data Augmentation: }
Since we use both camera data and LiDAR point clouds together, careful coordination between the camera and LiDAR data is necessary for data augmentation. We considered random flipping, rotation, scaling, and ground truth boxes sampling augmentation (GT-AUG) \cite{second}. We randomly flipped the LiDAR points and rotate the point clouds within a range of [$-\tfrac{\pi}{4}, \tfrac{\pi}{4}$] along the $z$ axis. We also scaled the coordinates of the points with a factor within $[0.95, 1.05]$. The modifications applied to the LiDAR points were reflected in the camera images. 
However, it was difficult to apply GT-AUG to both LiDAR and camera data without distortion. Hence, GT-AUG was used only when the LiDAR backbone network was pretrained.
We found that the benefit of the GT-AUG was not negligible in KITTI due to relatively small dataset size. 

{\bf Results on KITTI Test Set: }
Table \ref{table:test} provides the mAP performance of several 3D object detectors evaluated on KITTI 3D object detection tasks. The results for other algorithms are brought from the KITTI leaderboard  (\url{http://www.cvlibs.net/datasets/kitti/eval_object.php?obj_benchmark=3d}). We observe that the proposed 3D-CVF achieves the significant performance gain over other camera-LiDAR fusion-based detectors in the leaderboard. In particular, the 3D-CVF achieves up to 2.89\% gains (for hard difficulty) over UberATG-MMF \cite{mmf}, the best fusion-based method so far. The 3D-CVF outperforms most of the LiDAR-based 3D object detectors except for the STD \cite{std}. While the 3D-CVF outperforms the STD \cite{std} for easy and moderate levels but it is not for the hard level. Since the STD \cite{std} uses  the PointNet-based backbone, it might have a stronger LiDAR pipeline than the voxel-based backbone used in our 3D-CVF. It would be possible to apply our sensor fusion strategies to these kinds of detectors to improve their performance. 

Table \ref{table:test} also provides the inference time of 3D object detectors. We evaluated the interference time on 1 $\times$ NVIDIA GTX 1080 Ti. Note that the inference time of the proposed 3D-CVF is 75ms per frame, which looks comparable to that of other methods. We also measured the runtime of our LiDAR-only baseline. Note that the camera-LiDAR fusion  requires only 25ms additional runtime over 50ms runtime of the LiDAR-only baseline. 

\input{./table/ns_test.tex}

\subsection{nuScenes}
\hspace{\parindent}The nuScenes dataset is a large-scale 3D detection dataset that contains more than 1,000 scenes in Boston and Singapore \cite{nuScenes}. The dataset was collected using six multi-view cameras and 32-channel LiDAR.  360-degree object annotations for 10 object classes were provided. 
The dataset consists of 28,130 training samples and 6,019 validation samples. The nuScenes dataset suggests the use of an evaluation metric called nuScenes detection score (NDS)  \cite{nuScenes}.

{\bf Training Configuration: }
For the nuScenes dataset, the range of point cloud was within [$-49.6,-49.6$] $\times$[$-49.6,49.6$]$\times$[$-5,3$]$m$ in $(x,y,z)$ axis which was voxelized with each voxel size of $0.05\times0.05\times0.2m$. Consequently, this partitioning leads to  the voxel structure of size $1984\times1984\times40$. Anchor size of each class was determined by averaging the width and height values of the ground truths. We trained the network over 20 epochs using the same learning rate scheduling used in the KITTI dataset. The mini-batch size was set to 6.  DS sampling \cite{megvil} was adopted to alleviate the class imbalance problem in the nuScenes dataset.

{\bf Data Augmentation: }
For data augmentation, we used the same augmentation strategies except for GT-AUG. Unlike KITTI dataset, we found that skipping GT-AUG does not degrade the accuracy in nuScenes dataset.

{\bf Results on nuScenes Validation Set: }
We mainly tested our 3D-CVF on nuScenes to verify the performance gain achieved by sensor fusion. For this purpose, we compared the proposed 3D-CVF with the baseline algorithm, which has the same structure as our method except that the camera pipeline is disabled.  For a fair comparison, DS sampling strategy was also applied to the baseline. As a reference, we also added the performance of the SECOND \cite{second}, PointPillar \cite{pointpillar}, and MEGVII \cite{megvil}.  Table \ref{table:ns_test} provides the AP for 8 classes, mAP, and NDS achieved by several 3D object detectors.
We observe that the sensor fusion offers 2.74\%  and 3.57\% performance gains over the baseline in the mAP and NDS metrics,  respectively. The performance of the proposed method consistently outperforms the baseline in terms of AP for all classes. In particular, the detection accuracy is significantly improved for classes with
relatively low APs. This shows that the camera modality is useful to detect objects that are relatively difficult to identify with LiDAR sensors.

\subsection{Ablation study}
\hspace{\parindent}In Table \ref{table:ablation_new}, we present an ablation study for validating the effect of the ideas in the proposed 3D-CVF method. Note that our ablation study has been conducted on the KITTI {\it valid} set. Overall, our ablation study shows that the fusion strategy used in our 3D-CVF offers 1.32\%, 1.57\%, and 1.39\% gains in AP$_{Easy}$ AP$_{Mod.}$ and AP$_{Hard}$ over the LiDAR-only baseline. 

\input{./table/ablation_new.tex}

{\bf Effect of Naive Camera-LiDAR fusion: } We observe that when  the camera and LiDAR features are fused without cross-view feature mapping, adaptive gated fusion network, and 3D RoI fusion-based refinement, the improvement in detection accuracy is marginal. 

{\bf Effect of Adaptive Gated Fusion Network: } The adaptive gated fusion network leads to 0.54\%, 0.87\%, and 0.79\% performance boost in AP$_{Easy}$, AP$_{Mod.}$ and AP$_{Hard}$ levels, respectively. By combining the camera and LiDAR features selectively depending on their relevance to the detection task, our method can generate the enhanced joint camera-LiDAR feature. 

{\bf Effect of Cross-View Feature Mapping: } The auto-calibrated projection generates the smooth and dense camera features in the BEV domain. 
The detection accuracy improves over the baseline by 0.5\%, 0.06\%, and 0.15\%  in AP$_{Easy}$ AP$_{Mod.}$ and AP$_{Hard}$, respectively. 

{\bf Effect of 3D RoI Fusion-based Refinement: }
We observe that the 3D RoI fusion-based refinement improves AP$_{Easy}$ AP$_{Mod.}$ and AP$_{Hard}$ by 0.28\%, 0.63\%, and 0.45\%, respectively. It indicates that our 3D RoI fusion-based refinement compensates the lack of spatial information in the joint camera-LiDAR features  that may occur due to processing through many CNN pipelines.

\subsection{Performance Evaluation based on Object Distance}
\hspace{\parindent}To investigate the effectiveness of sensor fusion, we evaluated the detection accuracy of the 3D-CVF for different object distances. We categorized the objects in the KITTI {\it valid} set into three classes according to the distance ranges (0$\sim$20$m$), (20$\sim$40$m$), and (40$\sim$70$m$).  Table \ref{table:distance} provides the mAPs achieved by the 3D-CVF for three classes of objects. Note that the performance gain achieved by the sensor fusion is significantly higher for distant objects. The difference of mAP between nearby and distant objects is up to 5\%. This result indicates that the LiDAR-only baseline is not sufficient to detect distant objects due to the sparseness of LiDAR points and the camera modality  successfully compensates it.

\input{./table/distance_2.tex}

\section{Conclusions}
\hspace{\parindent}In this paper, we proposed a new camera and LiDAR fusion architecture for 3D object detection. The 3D-CVF achieved multi-modal fusion over two object detection stages. In the first stage, to generate the effective joint representation of camera and LiDAR data, we introduced the cross-view feature mapping that transforms the camera-view feature map into the calibrated and interpolated feature map in BEV. The camera and LiDAR features were selectively combined based on the relevance to the detection task using the adaptive gated fusion network. 
In the second stage, the 3D RoI-based fusion network refined the region proposals by pooling low-level camera and LiDAR features by 3D RoI pooling and fusing them after PointNet encoding. 
Our evaluation conducted on KITTI and nuScenes datasets confirmed that significant performance gain was achieved by the camera-LiDAR fusion and the proposed 3D-CVF outperformed the state-of-the-art 3D object detectors in KITTI leaderboard.

\section*{Acknowledgements}
\hspace{\parindent}This work was supported by Institute of Information \& Communications Technology Planning \& Evaluation (IITP) grant funded by the Korea government (MSIT) (2016-0-00564, Development of Intelligent Interaction Technology Based on Context Awareness and Human Intention Understanding).

\bibliographystyle{splncs04}
\bibliography{egbib}
\end{document}

%% file: table/test_3d.tex
\newcolumntype{C}{>{\centering\arraybackslash}p{4em}}
\newcolumntype{M}{>{\centering\arraybackslash}p{7em}}
\newcolumntype{K}{>{\centering\arraybackslash}p{4em}}
\renewcommand{\arraystretch}{1.1}
\begin{table*}[t]
\begin{center}
\begin{adjustbox}{width=0.75\textwidth}
\begin{tabular}{c|M|K||C|C|C}
\Xhline{4\arrayrulewidth}
\multirow{2}*{\it Method} &\multirow{2}*{\it Modality} & {\it Runtime} & \multicolumn{3}{c}{3D AP (\%)} \\ \cline{4-6}
&&{\it (ms)}&AP$_{Easy}$&AP$_{Mod.}$&AP$_{Hard}$\\
\hline\hline
VoxelNet \cite{voxelnet} & LiDAR &220& 77.47 & 65.11 & 57.73  \\
SECOND \cite{second} & LiDAR  &50& 83.13 & 73.66 & 66.20 \\
PointPillars \cite{pointpillar}& LiDAR &16.2& 79.05 & 74.99 & 68.30\\
PointRCNN \cite{pointrcnn}& LiDAR &100& 85.94 & 75.76 & 68.32 \\
Fast PointRCNN \cite{fpointrcnn} & LiDAR &65& 85.29 & 77.40 & 70.24 \\
Patches \cite{patches} & LiDAR &150& 88.67 & 77.20 & 71.82  \\
Part A$^2$ \cite{parta2}& LiDAR &80& 87.81 & 78.49 & 73.51\\
STD \cite{std}& LiDAR   &80& 87.95 & 79.71 & {\bf 75.09} \\
\hline
MV3D \cite{mmf} & LiDAR+RGB &240& 71.09 & 62.35 & 55.12 \\
AVOD \cite{avod} & LiDAR+RGB &80& 73.59 & 65.78 & 58.38 \\
F-PointNet \cite{fpointnet}& LiDAR+RGB &170& 81.20 & 70.39&62.19\\
AVOD-FPN \cite{avod}& LiDAR+RGB &100& 81.94 & 71.88 & 66.38\\
UberATG-ContFuse \cite{contfuse}& LiDAR+RGB &60& 82.54 & 66.22 & 64.04\\
RoarNet \cite{roarnet}& LiDAR+RGB &100& 83.95 & 75.79 & 67.88\\
UberATG-MMF \cite{mmf}& LiDAR+RGB &80& 88.40 & 77.43 & 70.22\\
\hline
Our 3D-CVF & LiDAR+RGB &75 &  {\bf 89.20} & {\bf 80.05} & 73.11 \\
\Xhline{4\arrayrulewidth}
\end{tabular}
\end{adjustbox}
\end{center}
\caption{\textbf{Performance on KITTI test benchmark for Car category:} The model is trained on KITTI training set and evaluated on KITTI test set.  ``AP$_{Easy}$",``AP$_{Mod.}$", and ``AP$_{Hard}$" mean the average precision for ``easy", ``moderate", and ``hard" difficulty levels.}
\label{table:test}
\end{table*}
\renewcommand{\arraystretch}{1}

%% file: table/ns_test.tex
\newcolumntype{C}{>{\centering\arraybackslash}p{3.5em}}
\renewcommand{\arraystretch}{1.1}
\begin{table*}[t]
\begin{center}
\begin{adjustbox}{width=1.0\textwidth}
\begin{tabular}{c||C|C|C|C|C|C|C|C||C|C}
\Xhline{5\arrayrulewidth}
             & Car & Ped. & Bus & Barrier & T.C. & Truck & Trailer & Moto.  & mAP & NDS \\\hline\hline
SECOND \cite{second} & 69.16 & 58.60 & 34.87 & 28.94 &24.83 & 23.73 & 5.52 & 16.60  & 26.32 & 35.36  \\
PointPillars \cite{pointpillar}& 75.25 & 59.47 & 43.80 & 30.95 & 18.57 & 23.42 & 20.15 & 21.12   &   29.34 & 39.03 \\
MEGVII \cite{megvil}           & 71.61 & 65.28 & 50.29 & {\bf 48.62} & {\bf 45.65} & 35.77 & 20.19 & 28.20   &37.68    & 44.15 \\\hline

LiDAR-only Baseline & 78.21 & 68.72 & 51.02 & 43.42 & 37.47 & 34.84 & 32.01 & 34.55   &   39.43  & 46.21\\
Our 3D-CVF & {\bf 79.69} & {\bf 71.28} & {\bf 54.96} & 47.10 & 40.82 & {\bf 37.94} & {\bf 36.29} & {\bf 37.18} & {\bf 42.17} & {\bf 49.78} \\ \Xhline{5\arrayrulewidth}
\end{tabular}
\end{adjustbox}
\end{center}
\caption{\textbf{mAP and NDS performance on nuScenes validation set:} The model was trained on nuScenes train set and evaluated on nuScenes validation set.  ``Cons. Veh." and ``Bicycle" classes were omitted as their accuracy was too low. The performance of the SECOND, PointPillars, and MEGVII was reproduced using their official codes.}
\label{table:ns_test}
\end{table*}
\renewcommand{\arraystretch}{1}

%% file: table/ablation_new.tex
\newcolumntype{C}{>{\centering\arraybackslash}p{6.3em}}
\newcolumntype{M}{>{\centering\arraybackslash}p{4.2em}}
\newcolumntype{A}{>{\centering\arraybackslash}p{3.8em}}
\renewcommand{\arraystretch}{1.2}
\begin{table}[t]
\begin{center}
\begin{adjustbox}{width=1.0\textwidth}
\begin{tabular}{c||M|C|C|C||A|A|A}
\Xhline{4\arrayrulewidth}
 \multirow{3}{*}{{\it Method}}& \multirow{3}{*}{\it Modality} & \multicolumn{3}{c||}{{\it Proposed Fusion Strategy}} &\multicolumn{3}{c}{3D AP (\%)}\\\cline{3-8}
 & &\multirow{2}{*}{\begin{tabular}[c]{@{}c@{}}Adaptive\\Gated Fusion\end{tabular}}& \multirow{2}{*}{\begin{tabular}[c]{@{}c@{}}Cross-View\\ Mapping\end{tabular}} & \multirow{2}{*}{\begin{tabular}[c]{@{}c@{}}3D RoI-based\\Refinement\end{tabular}}& \multirow{2}{*}{AP$_{Easy}$}& \multirow{2}{*}{AP$_{Mod.}$}& \multirow{2}{*}{AP$_{Hard}$}\\ &&&&&&&\\
 \hline\hline
LiDAR-only Baseline &LiDAR & & &      & 88.35 & 78.31 & 77.08  \\
\hline 
\multirow{4}{*}{Our 3D-CVF} &\multirow{4}{*}{\begin{tabular}[c]{@{}c@{}}LiDAR\\ + \\RGB\end{tabular}} & & & & 88.74 & 78.54 & 77.25   \\
&& \checkmark &  & & 88.89 & 79.19 & 77.87  \\
&& \checkmark & \checkmark  & & 89.39 & 79.25 &78.02  \\
&& \checkmark & \checkmark & \checkmark & {\bf 89.67} & {\bf 79.88} & {\bf 78.47}  \\

\Xhline{4\arrayrulewidth}
\end{tabular}
\end{adjustbox}
\end{center}
\caption{{\bf Ablation study on KITTI {\it valid} set for Car category: } The effect of our camera-LiDAR fusion schemes is highlighted in this study.}
\label{table:ablation_new}
\end{table}
\renewcommand{\arraystretch}{1}

%% file: table/distance_2.tex
\newcolumntype{C}{>{\centering\arraybackslash}p{6.3em}}
\renewcommand{\arraystretch}{1.4}
\begin{table*}[t]
\begin{center}
\begin{adjustbox}{width=0.65\textwidth}
\begin{tabular}{c||C|C|C}
\Xhline{3\arrayrulewidth}
            \multirow{2}{*}{\it Method}& \multicolumn{3}{c}{3D AP (\%)} \\\cline{2-4}
            &  $0\sim20m$    & $20\sim40m$  & $40\sim70m$   \\\cline{2-4}
\hline\hline
LiDAR-only Baseline& 89.86   &   76.72      &     30.57         \\\hline
Our 3D-CVF& {\bf 90.02} &    {\bf 79.73}    &  \textbf{35.86}       \\
{\it improvement} & {\it +0.16} & {\it +3.01} &  {\it +5.29} \\
\Xhline{4\arrayrulewidth}
\end{tabular}
\end{adjustbox}
\end{center}
\caption{\textbf{Accuracy of 3D-CVF for different object distance ranges: } The model is trained on KITTI {\it train} set and evaluated on KITTI {\it valid} set. We provide  the  detection accuracy  of  the  3D-CVF  for  object distance ranges, (0$\sim$20m), (20$\sim$40m), and (40$\sim$70m).}
\label{table:distance}
\vspace{-0.5cm}
\end{table*}

%% file: 5913.bbl
\begin{thebibliography}{10}
\providecommand{\url}[1]{\texttt{#1}}
\providecommand{\urlprefix}{URL }
\providecommand{\doi}[1]{https://doi.org/#1}

\bibitem{nuScenes}
Caesar, H., Bankiti, V., Lang, A.H., Vora, S., Liong, V.E., Xu, Q., Krishnan,
  A., Pan, Y., Baldan, G., Beijbom, O.: nuscenes: A multimodal dataset for
  autonomous driving. arXiv preprint arXiv:1903.11027  (2019)

\bibitem{mv3d}
Chen, X., Ma, H., Wan, J., Li, B., Xia, T.: Multi-view 3d object detection
  network for autonomous driving. In: Proceedings of the IEEE conference on
  Computer Vision and Pattern Recognition (CVPR). pp. 1907--1915 (2017)

\bibitem{fpointrcnn}
Chen, Y., Liu, S., Shen, X., Jia, J.: Fast point r-cnn. In: Proceedings of the
  IEEE International Conference on Computer Vision (ICCV). pp. 9775--9784
  (2019)

\bibitem{kitti}
Geiger, A., Lenz, P., Urtasun, R.: Are we ready for autonomous driving? the
  kitti vision benchmark suite. In: Proceedings of the IEEE conference on
  Computer Vision and Pattern Recognition (CVPR). pp. 3354--3361. IEEE (2012)

\bibitem{fastrcnn}
Girshick, R.: Fast r-cnn. IEEE International Conference on Computer Vision
  (ICCV) pp. 1440--1448 (2015)

\bibitem{resnet}
He, K., Zhang, X., Ren, S., Sun, J.: Deep residual learning for image
  recognition. In: Proceedings of the IEEE conference on Computer Vision and
  Pattern Recognition (CVPR). pp. 770--778 (2016)

\bibitem{jkkim_accv}
Kim, J., Koh, J., Kim, Y., Choi, J., Hwang, Y., Choi, J.W.: Robust deep
  multi-modal learning based on gated information fusion network. In: Asian
  Conference on Computer Vision (ACCV). pp. 90--106. Springer (2018)

\bibitem{avod}
Ku, J., Mozifian, M., Lee, J., Harakeh, A., Waslander, S.L.: Joint 3d proposal
  generation and object detection from view aggregation. In: IEEE/RSJ
  International Conference on Intelligent Robots and Systems (IROS). pp.~1--8.
  IEEE (2018)

\bibitem{pointpillar}
Lang, A.H., Vora, S., Caesar, H., Zhou, L., Yang, J., Beijbom, O.:
  Pointpillars: Fast encoders for object detection from point clouds. In:
  Proceedings of the IEEE conference on Computer Vision and Pattern Recognition
  (CVPR). pp. 12697--12705 (2019)

\bibitem{patches}
Lehner, J., Mitterecker, A., Adler, T., Hofmarcher, M., Nessler, B.,
  Hochreiter, S.: Patch refinement--localized 3d object detection. arXiv
  preprint arXiv:1910.04093  (2019)

\bibitem{Gs3d}
Li, B., Ouyang, W., Sheng, L., Zeng, X., Wang, X.: Gs3d: An efficient 3d object
  detection framework for autonomous driving. In: Proceedings of the IEEE
  Conference on Computer Vision and Pattern Recognition. pp. 1019--1028 (2019)

\bibitem{mmf}
Liang, M., Yang, B., Chen, Y., Hu, R., Urtasun, R.: Multi-task multi-sensor
  fusion for 3d object detection. In: Proceedings of the IEEE conference on
  Computer Vision and Pattern Recognition (CVPR). pp. 7345--7353 (2019)

\bibitem{contfuse}
Liang, M., Yang, B., Wang, S., Urtasun, R.: Deep continuous fusion for
  multi-sensor 3d object detection. In: Proceedings of the European Conference
  on Computer Vision (ECCV). pp. 641--656 (2018)

\bibitem{fpn}
Lin, T.Y., Doll{\'a}r, P., Girshick, R., He, K., Hariharan, B., Belongie, S.:
  Feature pyramid networks for object detection. In: Proceedings of the IEEE
  conference on Computer Vision and Pattern Recognition (CVPR). pp. 2117--2125
  (2017)

\bibitem{focal}
Lin, T.Y., Goyal, P., Girshick, R., He, K., Doll{\'a}r, P.: Focal loss for
  dense object detection. In: Proceedings of the IEEE International Conference
  on Computer Vision (ICCV). pp. 2980--2988 (2017)

\bibitem{ssd}
Liu, W., Anguelov, D., Erhan, D., Szegedy, C., Reed, S., Fu, C.Y., Berg, A.C.:
  Ssd: Single shot multibox detector. European Conference on Computer Vision
  (ECCV) pp. 21--37 (2016)

\bibitem{fpointnet}
Qi, C.R., Liu, W., Wu, C., Su, H., Guibas, L.J.: Frustum pointnets for 3d
  object detection from rgb-d data. In: Proceedings of the IEEE conference on
  Computer Vision and Pattern Recognition (CVPR). pp. 918--927 (2018)

\bibitem{pointnet}
Qi, C.R., Su, H., Mo, K., Guibas, L.J.: Pointnet: Deep learning on point sets
  for 3d classification and segmentation. In: Proceedings of the IEEE
  conference on Computer Vision and Pattern Recognition (CVPR). pp. 652--660
  (2017)

\bibitem{pointnet++}
Qi, C.R., Yi, L., Su, H., Guibas, L.J.: Pointnet++: Deep hierarchical feature
  learning on point sets in a metric space. In: Advances in Neural Information
  Processing Systems (NeurIPS). pp. 5099--5108 (2017)

\bibitem{yolo}
Redmon, J., Farhadi, A.: Yolo9000: Better, faster, stronger. Proceedings of the
  IEEE conference on Computer Vision and Pattern Recognition (CVPR) pp.
  6517--6525 (2017)

\bibitem{fasterrcnn}
Ren, S., He, K., Girshick, R., Sun, J.: Faster r-cnn: Towards real-time object
  detection with region proposal networks. Advances in Neural Information
  Processing Systems (NeurIPS) pp. 91--99 (2015)

\bibitem{pointrcnn}
Shi, S., Wang, X., Li, H.: Pointrcnn: 3d object proposal generation and
  detection from point cloud. In: Proceedings of the IEEE conference on
  Computer Vision and Pattern Recognition (CVPR). pp. 770--779 (2019)

\bibitem{parta2}
Shi, S., Wang, Z., Wang, X., Li, H.: Part-a\^{} 2 net: 3d part-aware and
  aggregation neural network for object detection from point cloud. arXiv
  preprint arXiv:1907.03670  (2019)

\bibitem{roarnet}
Shin, K., Kwon, Y.P., Tomizuka, M.: Roarnet: A robust 3d object detection based
  on region approximation refinement. In: IEEE Intelligent Vehicles Symposium
  (IV). pp. 2510--2515. IEEE (2019)

\bibitem{1cycle}
Smith, L.N.: A disciplined approach to neural network hyper-parameters: Part
  1--learning rate, batch size, momentum, and weight decay. arXiv preprint
  arXiv:1803.09820  (2018)

\bibitem{fconvnet}
Wang, Z., Jia, K.: Frustum convnet: Sliding frustums to aggregate local
  point-wise features for amodal 3d object detection. arXiv preprint
  arXiv:1903.01864  (2019)

\bibitem{pointfusion}
Xu, D., Anguelov, D., Jain, A.: Pointfusion: Deep sensor fusion for 3d bounding
  box estimation. In: Proceedings of the IEEE conference on Computer Vision and
  Pattern Recognition (CVPR). pp. 244--253 (2018)

\bibitem{second}
Yan, Y., Mao, Y., Li, B.: Second: Sparsely embedded convolutional detection.
  Sensors  \textbf{18}(10), ~3337 (2018)

\bibitem{pixor}
Yang, B., Luo, W., Urtasun, R.: Pixor: Real-time 3d object detection from point
  clouds. In: Proceedings of the IEEE conference on Computer Vision and Pattern
  Recognition (CVPR). pp. 7652--7660 (2018)

\bibitem{std}
Yang, Z., Sun, Y., Liu, S., Shen, X., Jia, J.: Std: Sparse-to-dense 3d object
  detector for point cloud. In: Proceedings of the IEEE International
  Conference on Computer Vision (ICCV). pp. 1951--1960 (2019)

\bibitem{voxelnet}
Zhou, Y., Tuzel, O.: Voxelnet: End-to-end learning for point cloud based 3d
  object detection. In: Proceedings of the IEEE conference on Computer Vision
  and Pattern Recognition (CVPR). pp. 4490--4499 (2018)

\bibitem{megvil}
Zhu, B., Jiang, Z., Zhou, X., Li, Z., Yu, G.: Class-balanced grouping and
  sampling for point cloud 3d object detection. arXiv preprint arXiv:1908.09492
   (2019)

\end{thebibliography}
